\pdfoutput=1

\documentclass[11pt]{article}

\usepackage[preprint]{acl}

\usepackage{times}
\usepackage{latexsym}

\usepackage[T1]{fontenc}

\usepackage[utf8]{inputenc}

\usepackage{microtype}

\usepackage{inconsolata}

\usepackage{graphicx}

\usepackage{booktabs}
\usepackage{multirow}
\usepackage{xcolor}
\usepackage{colortbl}
\usepackage{tikz}

\definecolor{keywords}{RGB}{255,0,90}
\definecolor{comments}{RGB}{0,0,113}
\definecolor{red}{RGB}{160,0,0}
\definecolor{green}{RGB}{0,150,0}

\definecolor{bananayellow}{rgb}{1.0, 0.88, 0.21}

\usepackage{hyperref}
\usepackage{amsmath}

\usepackage{subcaption}
\usepackage{enumitem}
\usepackage{bbding}
\usepackage{amssymb}

\usepackage[linesnumbered,lined,boxed,commentsnumbered]{algorithm2e}
\usepackage{array}

\title{Ask-Before-Detection: Identifying and Mitigating Conformity Bias in LLM-Powered Error Detector for Math Word Problem Solutions}

\author{
 \textbf{Hang Li\textsuperscript{1}},
 \textbf{Tianlong Xu\textsuperscript{2}},
 \textbf{Kaiqi Yang\textsuperscript{1}},
 \textbf{Yucheng Chu\textsuperscript{1}},
\\
 \textbf{Yanling Chen\textsuperscript{1}},
 \textbf{Yichi Song\textsuperscript{3}},
 \textbf{Qingsong Wen\textsuperscript{2}},
 \textbf{Hui Liu\textsuperscript{1}}\thanks{Corresponding author.}
\\
 \textsuperscript{1}Michigan State University,
 \textsuperscript{2}Squirrel Ai Learning,
 \textsuperscript{3}Carleton College,
\\
 \texttt{\{lihang4,kqyang,chuyuch2,chenya67,liuhui7\}@msu.edu} \\
 \texttt{\{tianlongxu,qingsongwen\}@squirrelai.com}, \texttt{\{songc2\}@carleton.edu}
}

\begin{document}
\maketitle
\begin{abstract}

The rise of large language models (LLMs) offers new opportunities for automatic error detection in education, particularly for math word problems (MWPs). While prior studies demonstrate the promise of LLMs as error detectors, they overlook the presence of multiple valid solutions for a single MWP. Our preliminary analysis reveals a significant performance gap between conventional and alternative solutions in MWPs, a phenomenon we term conformity bias in this work. To mitigate this bias, we introduce the Ask-Before-Detect (AskBD) framework, which generates adaptive reference solutions using LLMs to enhance error detection. Experiments on 200 examples of GSM8K show that AskBD effectively mitigates bias and improves performance, especially when combined with reasoning-enhancing techniques like chain-of-thought prompting.

\end{abstract}

\section{Introduction}
\vspace{-0.2cm}
Automatic Error Detection (AED) has been a prominent research topic in education over the past few decades~\citep{leacock2014automated}. Supported by rapid advancements in natural language processing (NLP) technologies, particularly in language modeling~\citep{min2023recent}, AED research has achieved notable success in language education~\citep{huang2023trends}. The recent emergence of large language models (LLMs) presents new opportunities for AED studies. Leveraging their exceptional capabilities in logical reasoning~\citep{pan2023logic}, LLMs have become promising tools helping the quick development of AED in more challenging scenarios including programming~\citep{messer2024automated} and mathematics learning~\citep{jiang2024llms}. Recent studies have introduced benchmark datasets to demonstrate the potential of LLMs in AED across diverse domains~\citep{yan2024errorradar}. Moreover, due to the reasoning-intensive nature of error detection in mathematical problems, recent research has employed AED tasks on math word problems (MWPs) to evaluate the comparative reasoning capabilities of LLMs~\citep{li2024evaluating}. Studies~\citep{zhou2024your} have indicated that identifying errors in MWPs, rather than generating correct solutions, serves as a more effective metric for assessing differences in the reasoning capabilities of LLMs. In this paper, we explore AED for MWPs. Specifically, building on prior studies, we define the AED task as identifying both the erroneous step and its error category from a given input pair consisting of a question and its solution. It is important to note that a correct result requires accurate identification of both the error step and the error category. 

\begin{figure*}
    \centering
    \includegraphics[width=0.95\linewidth]{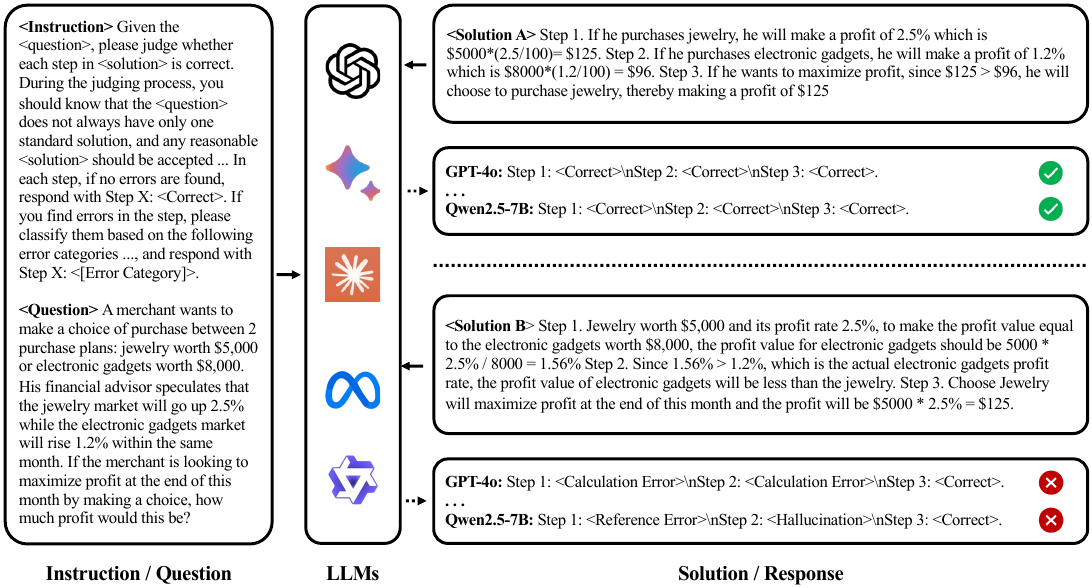}
    \vspace{-0.2cm}
    \caption{An illustration of error detection in MWP solutions: <Solution A> represents the conventional solution, which achieves accurate error detection with LLM-powered error detectors. In contrast, <Solution B>, while also correct, encounters erroneous error detection results across all LLM-powered error detectors.}
    \label{fig:sample_case}
    \vspace{-0.5cm}
\end{figure*}

While previous studies~\citep{li2024evaluating,zhou2024your} have explored various methods for evaluating the error detection capabilities of LLMs on MWPs, these approaches predominantly focus on generating erroneous solutions based solely on the conventional solutions provided in the dataset. In practice, however, a single MWP can have multiple valid solutions, leaving the performance of LLMs on alternative solutions largely unexplored. In Figure~\ref{fig:sample_case}, we present an illustrative example where both the conventional and alternative solutions are submitted to an LLM-powered error detector, yet only the conventional solution receives the correct detection result. Motivated by this observation, we propose an automatic method for generating alternative solutions and evaluate the behavior of LLM-powered error detectors on 200 pairs of conventional and alternative solutions. Our preliminary results in section~\ref{sec:bias_identification} reveal an average 7\% detection performance gap between conventional and alternative solutions, with even advanced closed-source models exhibiting the same limitation. These findings suggest that current LLM-based error detectors display a pronounced \textbf{conformity bias}, favoring a specific solution format while rejecting others. This bias is particularly concerning in educational contexts, as it discourages students from exploring diverse problem-solving approaches and stifles creativity.  

To investigate the underlying causes of conformity bias and develop effective strategies to mitigate it, we conduct further preliminary studies in Section~\ref{sec:grading_analysis}, which examines the common patterns in the behavior of LLM-powered error detectors when evaluating diverse solutions. Our findings reveal that error detection accuracy is closely correlated with the likelihood scores assigned by LLMs to solutions, with higher likelihood scores corresponding to improved detection performance. Since alternative solutions typically receive lower likelihood scores compared to conventional ones, conformity bias naturally emerges. This observation points to a potential remedy: adjusting the likelihood scores of solutions. However, this approach faces two significant challenges. First, fine-tuning advanced models requires high-quality datasets and incurs substantial costs. Second, while fine-tuning may improve likelihood scores for samples within the training dataset, its generalizability to novel solutions remains uncertain. During our investigation into the impact of introducing a reference answer during error detection, we observed that conformity bias is significantly reduced across all LLMs. This finding inspires us to leverage reference answers as a viable strategy for mitigating bias. However, uniformly providing a standard reference answer for every solution is suboptimal in practice. Misalignment between the reasoning behind different solutions and the reference answer can sometimes degrade final detection performance. To address this, we propose the Ask-Before-Detection (AskBD) framework, which generates adaptive reference answers through step-by-step question-answering techniques. By leveraging the strong problem-solving capabilities of LLMs and employing a decomposed, question-guided approach, the reference answers can be generated with high accuracy, even using less capable models. Incorporating these generated reference solutions significantly mitigates conformity bias in LLM-powered error detectors for MWP solutions. Furthermore, the adaptability of the generated references enhances the overall performance of LLM-powered error detector across both conventional and alternative solutions.

\vspace{-0.2cm}
\section{Preliminary Study}
\vspace{-0.2cm}
\label{sec:pre}
\begin{figure*}[!btph]
    \centering
    \includegraphics[width=0.95\linewidth]{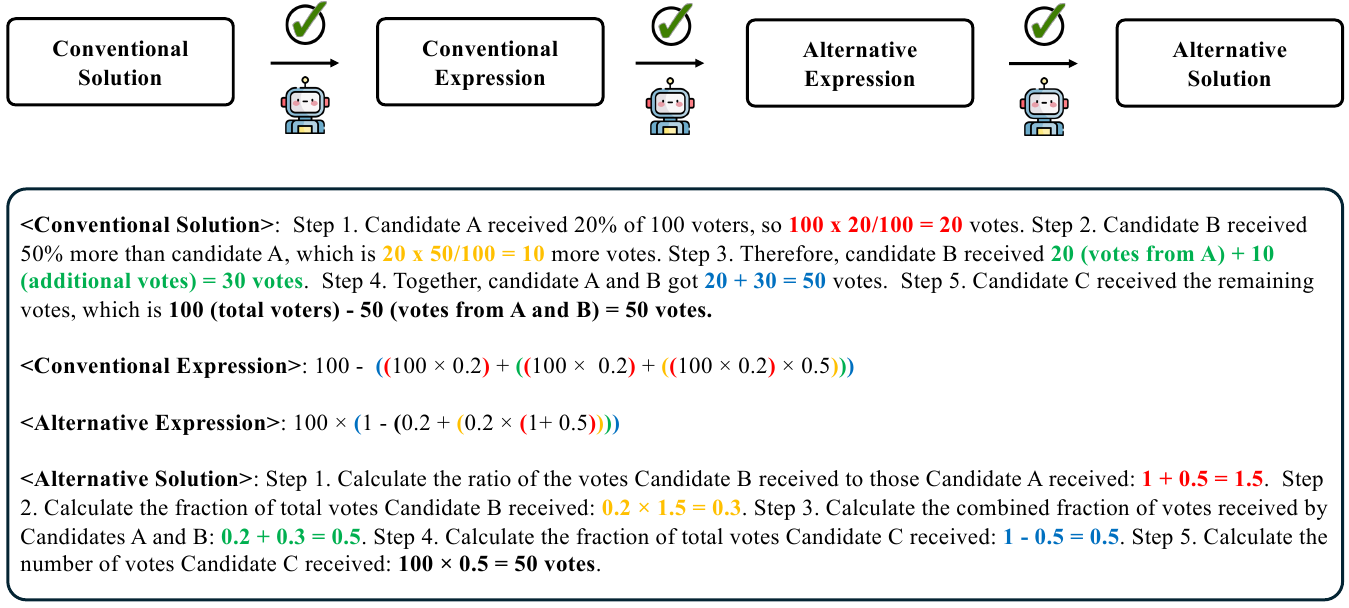}
    \vspace{-0.2cm}
    \caption{The ASP pipeline to generate permuted solution. The corresponding relationships between the calculations in each step and the parts enclosed by different parentheses in expression are highlighted using matching colors.}
    \label{fig:asp_pipeline}
    \vspace{-0.5cm}
\end{figure*}

In this section, we present our preliminary studies aimed at identifying and understanding the presence of \textbf{conformity bias} in LLM-powered error detectors. Specifically, we begin by describing our automated method for preparing an error detection dataset featured with paired conventional and alternative solutions. Then, we analyze the relationship between likelihood scores and the behavior patterns of LLM-powered error detectors. Finally, we present our observations on how incorporating reference solution text influences the performance and behavior of error detectors. 

\vspace{-0.2cm}
\subsection{Automatic Solution Permutation}
\label{sec:asp}
\vspace{-0.2cm}

Building a high-quality alternative solution dataset is critical to our preliminary study, as low-quality alternatives, such as simple semantic paraphrases of conventional solutions, fail to effectively expose the "conformity bias" in LLM-powered error detectors. During our initial exploration, we observed that directly using simple prompts to query LLMs for automatically generating alternative solutions presents significant challenges. Specifically, LLMs often produce paraphrased versions of conventional solutions unless detailed and specific instructions about the solving strategy are provided during the generation process. To address this, we propose the Automatic Solution Permutation (ASP) method, which leverages the correspondence between conventional solutions and their solving expressions. Using these expressions as specific instructions helps LLMs move beyond paraphrasing behavior, enabling the generation of high-quality alternative solutions.

The ASP method operates in three stages: Extract, Permute, and Explain. At each stage, LLMs are prompted to execute specific tasks independently. In the Extract stage, ASP encapsulates the steps of a conventional solution into a single mathematical expression. To ensure accuracy, these expressions are executed, and any that fail to produce correct answer values are discarded. In the Permute stage, ASP generates new expressions by applying operations such as factorization, distribution, and order rearrangement, which transform the expressions while preserving their mathematical equivalence. A similar filtering process is applied to these permuted expressions to ensure correctness. Finally, in the Explain stage, the permuted expressions are provided to LLMs to guide the generation of high-quality alternative solutions. By instructing the LLMs to interpret the brackets within each expression as distinct steps, the ASP method produces detailed, step-by-step alternative solutions. Figure~\ref{fig:asp_pipeline} illustrates the ASP pipeline and provides an example of paired conventional and alternative solutions alongside their corresponding solving expressions.

In our study, we use GPT-4o as the backbone LLM for each stage of the ASP method. To begin, we randomly sample 200 question-and-answer pairs from the test split of GSM8K~\citep{cobbe2021training} and use them to construct our conventional dataset, $\mathcal{D}$. For each sample in $\mathcal{D}$, we apply ASP three times to generate three candidate permuted solutions for each conventional solution. Subsequently, a graduate student from the education department reviews the quality of all the generated alternative solutions and selects the highest-quality alternative solution for each convention solution. These selected solutions are then compiled to form the alternative dataset, $\mathcal{D}'$.

\vspace{-0.2cm}
\subsection{Erroneous Solution Generation}
\label{sec:error_inject}

After preparing the alternative solution dataset, the next step is to generate erroneous solutions. Building on prior work~\citep{li2024evaluating}, which categorize common errors in solutions to MWPs into categories, we choose four representative error types that commonly encountered in real-world error grading scenarios: calculation errors ($\mathcal{E}_C$), reference errors ($\mathcal{E}_R$), missing steps ($\mathcal{E}_M$), and hallucinations ($\mathcal{E}_H$), for our study. It is worth noting that this study specifically aims to explore conformity bias, and therefore, we do not include all possible error types. To minimize the risk of experimental noise caused by ambiguous definitions, we defined these error types in a straightforward and easily distinguishable manner. Detailed descriptions of each error type are provided in Table~\ref{tab:error} in Appendix~\ref{apx:error_define}. To simulate erroneous solutions, we injected these errors into correct solutions using a generation strategy inspired by prior work~\cite{li2024evaluating}. During the injection process, the error type was controlled through a hyper-parameter, while the specific error location (error step number) was determined randomly. This approach enables controlled testing of the AED's ability to handle and identify various error scenarios effectively. For each example in $\mathcal{D}$ and $\mathcal{D}'$, we generated four corresponding erroneous solutions, each associated with one of the four error types. This process yielded a total of 2,000 examples, which were prepared for subsequent analysis.

\vspace{-0.2cm}
\subsection{Analysis and Findings}
\label{sec:grading_analysis}

Before delving into the details about our analysis and findings, we first introduce the evaluation metric used for our following analyses. Specifically, since the locations and categories of injected errors are automatically labeled during the error injection process for each solution, we task the LLM-powered error detector with identifying both the error locations and their types. The evaluation metric is the identification accuracy across both correct and erroneous solutions.

\subsubsection{Conformity Bias Identification}
\label{sec:bias_identification}

\begin{table}[]
\caption{Error detection performance on ordinary solutions ($\mathcal{D}$) and alternative solutions ($\mathcal{D}'$). The performance gap is calculated by $\Delta = \mathcal{D}' - \mathcal{D}$.}
\vspace{-0.2cm}
\label{tab:correct_reject}
\renewcommand{\arraystretch}{1.15}
\resizebox{0.48\textwidth}{!}{
\begin{tabular}{@{}c|ccc|ccc@{}}
\toprule
\multirow{2}{*}{\textbf{Model}} & \multicolumn{3}{c|}{\textbf{Base}} & \multicolumn{3}{c}{\textbf{Advance}} \\ \cmidrule(l){2-7} 
 & $\mathcal{D}$ & $\mathcal{D}'$ & $\Delta$ & $\mathcal{D}$ & $\mathcal{D}'$ & $\Delta$ \\ \midrule
GPT-4o & 27.2 & \multicolumn{1}{c|}{18.4} & {\color{red}\textbf{-8.8}} & 52.9 & \multicolumn{1}{c|}{43.4} & {\color{red}\textbf{-9.5}} \\
Claude-3.5 &  38.2 & \multicolumn{1}{c|}{34.7} & {\color{red}\textbf{-3.5}} & 59.9 & \multicolumn{1}{c|}{52.7} & {\color{red}\textbf{-7.2}} \\
Gemini-1.5 & 46.4 & \multicolumn{1}{c|}{39.5} & {\color{red}\textbf{-6.9}} & 65.2 & \multicolumn{1}{c|}{55.6} & {\color{red}\textbf{-9.6}} \\
Llama-3.1 &  20.2 & \multicolumn{1}{c|}{20.9} & {\color{green}\textbf{+0.7}} & 44.3 & \multicolumn{1}{c|}{37.2} & {\color{red}\textbf{-7.2}} \\
Qwen-2.5 & 24.7 & \multicolumn{1}{c|}{16.3} & {\color{red}\textbf{-8.4}} & 46.3 & \multicolumn{1}{c|}{38.8} & {\color{red}\textbf{-7.5}} \\ \bottomrule
\end{tabular}}
\vspace{-0.6cm}
\end{table}

To identify conformity bias, we employ a widely-used LLM-powered error detection approach, leveraging prompt engineering techniques outlined in previous studies~\citep{li2024evaluating}. In addition, the instruction text informs the LLMs that alternative solutions to the given question exist and emphasizes that all reasonable solutions should be accepted. To minimize variability in performance due to ambiguity in error categories, we provide explicit definitions for each error category within the prompt text, ensuring clarity for the LLMs. The prompt used for the error detection task is illustrated in Figure~\ref{fig:prompt_0}.

To comprehensively analyze the conformity bias exhibited by various LLMs, we conducted experiments with 10 representative models. These include three closed-source series with their advanced (base) versions (e.g., GPT-4o (Mini)~\cite{bubeck2023sparks}, Gemini-1.5-Pro (Flash)~\cite{team2023gemini}, Claude-3.5-Sonnet (Haiku)~\cite{TheC3}) and two open-source series with their advanced (base) counterparts (e.g., Llama-3.1-70B (8B)~\cite{touvron2023llama} and Qwen-2.5-72B (7B)~\cite{yang2024qwen2}). Table~\ref{tab:correct_reject} presents a comparison of the average error detection accuracy across both the conventional solution dataset $\mathcal{D}$ and the alternative solution dataset $\mathcal{D}'$. The results clearly demonstrate a consistent performance gap between the two datasets, confirming the presence of conformity bias in LLM-based error detection tasks.

\subsubsection{Solution Likelihood Score Analysis}

To investigate the underlying causes of conformity bias, we first chose to use the log-likelihood score, denoted as $\log L_\theta(s|q)$, returned by the LLM for a given solution text \(s\) to the question text \(q\), as an indicator. This likelihood score is utilized as it reflects the LLM's confidence in the solution text relative to the question text. If a solution known to be correct receives a low confidence score from the LLM, it suggests that the LLM does not fully understand the solution. Conversely, if the correct solution receives a high score, it indicates that the LLM is proficient with the solution. The detailed calculation method is presented below:
\vspace{-0.3cm}
\begin{equation}
\vspace{-0.3cm}
\label{eq:raw_likelihood}
    \log L_\theta(s|q) = \sum_{i=1}^{|s|} \log L_\theta(s_i|[q,s_{1:(i-1)}])
\end{equation}

\noindent where $\theta$ represents the parameters of the LLM, $[\cdot,\cdot]$ is text concatenation, $s_i$ is the $i$-th token of the solution text. However, directly comparing the likelihood scores calculated by Equation~\ref{eq:raw_likelihood} for solutions of varying lengths is still problematic, as the likelihood score is inversely proportional to the length of $s$. In other words, shorter solutions with fewer tokens tend to have higher scores than longer ones. To address this issue, we finally adopt the average token log-likelihood score for our analysis in subsequent studies.

\vspace{-0.3cm}
\begin{equation}
\vspace{-0.3cm}
\label{eq:indicator}
    \log \bar{L}_\theta(s|q) = \frac{\log L_\theta(s|q)}{|s|}
\end{equation}

\begin{figure*}
     \centering
     \begin{subfigure}[b]{0.32\textwidth}
         \centering
         \includegraphics[width=\textwidth]{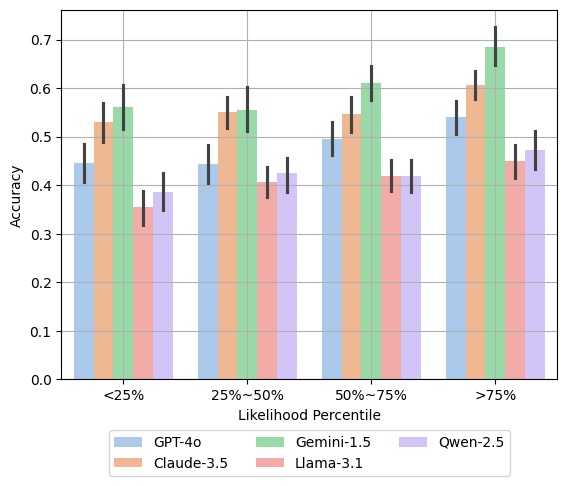}
         \caption{Advance models}
         \label{fig:like_adv}
     \end{subfigure}
     \begin{subfigure}[b]{0.32\textwidth}
         \centering
         \includegraphics[width=\textwidth]{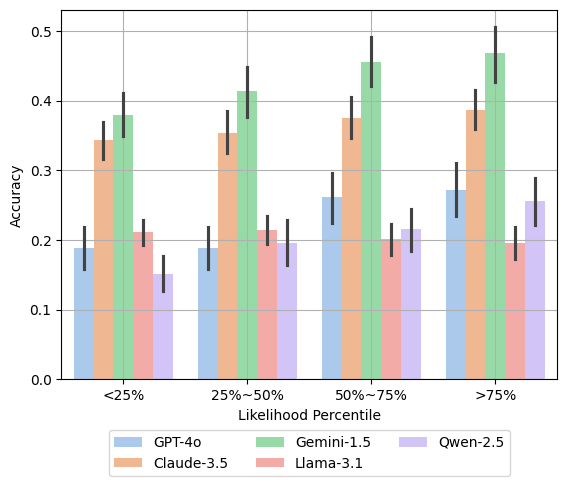}
         \caption{Base models}
         \label{fig:like_base}
     \end{subfigure}
     \begin{subfigure}[b]{0.32\textwidth}
         \centering
         \includegraphics[width=\textwidth]{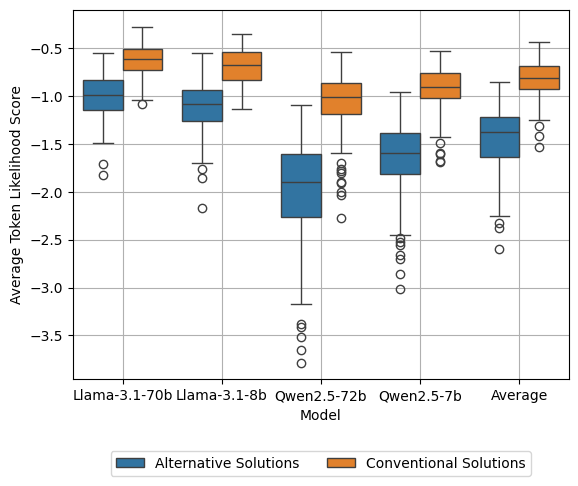}
         \caption{Likelihood distributions}
         \label{fig:like_dist}
     \end{subfigure}
     \vspace{-0.2cm}
     \caption{Average error detection accuracy across samples grouped by the 25th, 50th, and 75th percentiles of $I_u$.}
     \label{fig:likelihood}
     \vspace{-0.5cm}
\end{figure*}

In Figure~\ref{fig:like_adv} and Figure~\ref{fig:like_base}, we present the average error detection accuracy across different  likelihood score groups. Specifically, given the likelihood score to both convention and alternative solutions, we group them based on their likelihood score percentiles. For simplicity, we use the four quarters in our experiment. It is important to note that, since the likelihood scores of closed-source models are unavailable, we use the average scores of all open-source LLMs as a pseudo-indicator for this analysis. From the figure, we observe that the larger quarter groups with higher indicator values exhibit a clear advantage over those with the smaller quarter ones. In addition, we plot the likelihood score distribution comparisons between the solution from $\mathcal{D}$ and $\mathcal{D}'$ in Figure~\ref{fig:like_dist}. From these plot, we can draw a clear conclusion that the conformity bias in current LLM to error detection tasks is caused by its decreased understanding to those alternative solution. 

\subsubsection{Reference-based Detection Findings}
\label{sec:refer_detect}

Directly improving the likelihood scores of alternative solutions poses inherent challenges. Strategies like fine-tuning large language models (LLMs) primarily improve likelihood scores for training samples, but their effectiveness on unseen alternative solutions remains unpredictable. Building on prior work~\citep{daheim2024stepwise}, which demonstrated that introducing reference answers during error detection enhances performance on conventional solutions, we extend this approach to alternative solutions. It is important to note that, in real-world error detection scenarios, reference answers are not always available. Even when they are, conventional solutions are more commonly provided. Take this into consderation, we conducted experiments comparing two reference-based detection setups: (1) uniformly using conventional solutions as references and (2) adaptively using corresponding solutions as references. The detailed results are presented in Table~\ref{tab:mismatch_reference} and Table~\ref{tab:match_reference}, respectively. 

\begin{table}[]
\caption{Error detection performance w/ using corresponding solution as reference solution for both ordinary solutions ($\mathcal{D}$) and alternative solutions ($\mathcal{D}'$). The performance gap is calculated by $\Delta=\mathcal{D}'-\mathcal{D}$}
\label{tab:match_reference}
\vspace{-0.2cm}
\renewcommand{\arraystretch}{1.15}
\resizebox{0.48\textwidth}{!}{
\begin{tabular}{@{}c|ccc|ccc@{}}
\toprule
\multirow{2}{*}{\textbf{Model}} & \multicolumn{3}{c|}{\textbf{Base}} & \multicolumn{3}{c}{\textbf{Advance}} \\ \cmidrule(l){2-7} 
 & $\mathcal{D}$ & $\mathcal{D}'$ & $\Delta$ & $\mathcal{D}$ & $\mathcal{D}'$ & $\Delta$ \\ \midrule
GPT-4o & 60.0 & \multicolumn{1}{l|}{56.5} & {\color{red}\textbf{-3.5}} & 75.5 & \multicolumn{1}{l|}{73.8} & {\color{red}\textbf{-1.7}} \\
Claude-3.5 & 60.4 & \multicolumn{1}{l|}{57.9} & {\color{red}\textbf{-2.5}} & 84.0 & \multicolumn{1}{l|}{81.6} & {\color{red}\textbf{-2.4}} \\
Gemini-1.5 & 69.7 & \multicolumn{1}{l|}{66.7} & {\color{red}\textbf{-3.0}} & 85.3 & \multicolumn{1}{l|}{83.7} & {\color{red}\textbf{-1.6}} \\
Llama-3.1 & 34.6 & \multicolumn{1}{l|}{33.7} & {\color{red}\textbf{-0.9}} & 77.5 & \multicolumn{1}{l|}{79.4} & {\color{green}\textbf{+1.9}} \\
Qwen-2.5 & 54.4 & \multicolumn{1}{l|}{51.2} & {\color{red}\textbf{-3.2}} & 59.3 & \multicolumn{1}{l|}{60.8} & {\color{green}\textbf{+1.5}} \\ \bottomrule
\end{tabular}}
\vspace{-0.3cm}
\end{table}

\begin{table}[]
\caption{Error detection performance w/ using convention solution as reference for both ordinary solutions ($\mathcal{D}$) and alternative solutions ($\mathcal{D}'$). The performance gap is calculated by $\Delta = \mathcal{D}'-\mathcal{D}$}
\label{tab:mismatch_reference}
\vspace{-0.2cm}
\renewcommand{\arraystretch}{1.15}
\resizebox{0.48\textwidth}{!}{
\begin{tabular}{@{}c|ccc|ccc@{}}
\toprule
\multirow{2}{*}{\textbf{Model}} & \multicolumn{3}{c|}{\textbf{Base}} & \multicolumn{3}{c}{\textbf{Advance}} \\ \cmidrule(l){2-7} 
 & $\mathcal{D}$ & $\mathcal{D}'$ & $\Delta$ & $\mathcal{D}$ & $\mathcal{D}'$ & $\Delta$ \\ \midrule
GPT-4o & 60.0 & \multicolumn{1}{l|}{32.0} & {\color{red}\textbf{-28.0}} & 75.5 & \multicolumn{1}{l|}{53.3} & {\color{red}\textbf{-22.2}} \\
Claude-3.5 & 60.4 & \multicolumn{1}{l|}{38.9} & {\color{red}\textbf{-21.5}} & 84.0 & \multicolumn{1}{l|}{59.4} & {\color{red}\textbf{-24.6}} \\
Gemini-1.5 & 69.7 & \multicolumn{1}{l|}{50.8} & {\color{red}\textbf{-18.9}} & 85.3 & \multicolumn{1}{l|}{67.7} & {\color{red}\textbf{-17.6}} \\
Llama-3.1 & 34.6 & \multicolumn{1}{l|}{20.5} & {\color{red}\textbf{-14.1}} & 77.5 & \multicolumn{1}{l|}{48.8} & {\color{red}\textbf{-28.7}} \\
Qwen-2.5 & 54.4 & \multicolumn{1}{l|}{22.2} & {\color{red}\textbf{-32.2}} & 59.3 & \multicolumn{1}{l|}{43.5} & {\color{red}\textbf{-15.8}} \\ \bottomrule
\end{tabular}}
\vspace{-0.6cm}
\end{table}

By analyzing the results across Table~\ref{tab:correct_reject}, Table~\ref{tab:mismatch_reference}, and Table~\ref{tab:match_reference}, it is evident that introducing reference solutions improves error detection accuracy for both datasets, $\mathcal{D}$ and $\mathcal{D}'$. However, the choice of reference solution significantly impacts performance. While introducing corresponding reference solutions effectively mitigates bias, uniformly using conventional solutions tends to amplify it. This contrast highlights the critical importance of selecting appropriate reference solutions to enhance error detection in alternative scenarios.

\vspace{-0.1cm}
\section{Method}
\vspace{-0.1cm}
The findings in Section~\ref{sec:refer_detect} suggest that incorporating a reference solution during the detection process is an effective approach to addressing conformity bias. However, the choice of the reference solution plays a critical role. Building on this insight, we propose the Ask-Before-Detection (AskBD) framework, which leverages the generative capabilities of large language models (LLMs) to create adaptive reference solutions tailored to each provided solution during the grading process. The AskBD offers several advantages. First, it utilizes the inherent problem-solving capabilities of LLMs rather than relying on fine-tuning, which makes AskBD easily extendable to various solutions. Second, by adaptively generating reference solutions, the framework ensures that these references are well-aligned with the given answers, significantly reducing the risk of mismatches that could amplify bias. Furthermore, the AskBD is orthogonal to other reasoning techniques, such as chain-of-thought (CoT)~\citep{wei2022chain}, which can complement and enhance their performance. By integrating AskBD with these algorithms, the error detection capabilities of LLMs can be further improved. The overall structure of the AskBD is illustrated in Figure~\ref{fig:askbd} in Appendix~\ref{apx:naive_promot}. 

\begin{figure*}
    \centering
    \includegraphics[width=0.95\linewidth]{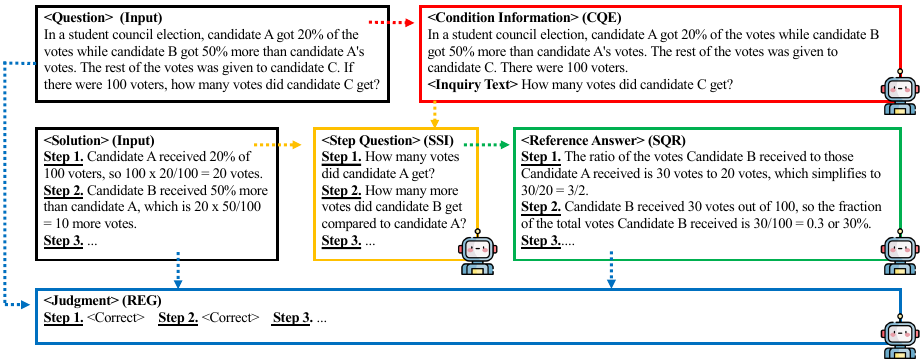}
    \vspace{-0.2cm}
    \caption{An overview of AskBG framework where steps are marked with colors. <Question> and <Solution> are the inputs and <Reference Answer> is generated by the framework, which used to generate the final response.}
    \label{fig:askbd}
    \vspace{-0.5cm}
\end{figure*}

\RestyleAlgo{ruled}

\vspace{-0.2cm}
\begin{algorithm}[!btph]
\caption{Ask-Before-Detection}\label{algo:askbd}
\KwIn{Question text $q$, solution text $s$, LLM $f_\theta$, prompt text $\mathcal{P}_{(\cdot)}$}
Condition and question extractor (CQE): Extract condition information $q_c$ and inquiry text $q_i$ from the question text $q$. $(q_c, q_i) = f_\theta([\mathcal{P}_{cqe},q])$;

Solution Step Inquirer (SSI): Convert solution text $s$ into step-wise question list text $Q$ and append inquiry text $q_i$ at the end. $Q = [f_\theta([\mathcal{P}_{ssi},s]), q_i]$;

Step Question Responder (SQR): Generate reference solution $r$ by summarizing the answers to each question in $Q$ using condition text $q_c$. $r = f_\theta([\mathcal{P}_{sqr},q_c,Q])$;

Reference-Enhanced Grader (REG): Generate the error detection result (error location $y_s$, error type $y_e$) based on the input ($q$, $s$, $r$). $y =f_\theta([\mathcal{P}_{reg},q,s,r])$;

\KwRet $y_{s},y_{e}$

\end{algorithm}
\vspace{-0.2cm}

AskBD consists of four components, executed sequentially to generate an adaptive reference answer tailored to the input solution.  First, the Condition and Question Extractor (CQE) processes the input question text, $q$, by extracting two key elements from the original question stem: condition information and inquiry text. The condition information, $q_c$, represents the known facts or context provided in the question, while the inquiry text, $q_i$, specifies the task or problem posed by the question. Then, the Solution Step Inquirer (SSI) focuses on generating step-specific questions based on the conclusions of each step in the provided solution, $s$. To improve the stability of the generated results, the SSI first summarizes the conclusion of each step before formulating corresponding questions. These step-specific questions are compiled into a question list, $Q$, with the inquiry text $q_i$ always appended to the end of the list to ensure that the original question’s task is addressed in the generated reference answer. Next, after both the condition text $q_c$ and the question list $Q$ are prepared, the Step Question Responder (SQR) generates responses to each question in $Q$ and reorganizes them into a referenced solution, $r$. Finally, the Reference-Enhanced Grader (REG) uses the referenced solution $r$ along with the inputs $q$ and $s$ to produce the final grading results. More details can be found in Algorithm~\ref{algo:askbd}.

\vspace{-0.2cm}
\section{Experiment}
\vspace{-0.2cm}
In this section, we present experiments to validate the effectiveness of AskBD. Experiments are designed to address the following research questions:
\vspace{-0.5cm}
\begin{itemize}[leftmargin=*]
    \item \textbf{RQ1:} Does AskBD help mitigate conformity bias in error detection?
    \vspace{-2mm}
    \item \textbf{RQ2:} What additional performance advantages does AskBD provide? 
    \vspace{-2mm}
    \item \textbf{RQ3:} How compatible is AskBD with other reasoning techniques, such as chain-of-thought?
\end{itemize}

\vspace{-0.2cm}
\subsection{Experimental Settings}

To answer above research questions, we use dataset generated during the preliminary study introduced in Section~\ref{sec:pre}. The detailed statistics of the dataset are shown as Table~\ref{tab:stats} in Appendix~\ref{ap:data_detail}. To evaluate the generalizability of AskBD, we implement it using the same 10 LLMs described in Section~\ref{sec:pre}, with detailed information about each model provided in Table~\ref{tab:llm_detail} in Appendix~\ref{apx:model_detail}. Additionally, we incorporate the CoT reasoning approach into the prompts to assess its compatibility with AskBD. Each experiment is conducted using three different random seeds, and we report the mean error detection accuracy in the results. As this is the first study to systematically examine the occurrence of conformity bias in LLM-powered error detection for MWP solutions, the results from the preliminary study sections serve as one baseline. Furthermore, CoT, being an orthogonal method to AskBD, is treated as another baseline for comparison.

\vspace{-0.1cm}
\subsection{Result and Discussion}

\begin{table*}[]
\centering
\caption{Error detection performance w/ different baseline methods ($\mathcal{M}_0$: Naive prompt, $\mathcal{M}_1$: CoT prompt, $\mathcal{M}_2$: Naive prompt + AskBD, $\mathcal{M}_3$: CoT prompt + AskBD) on ordinary solutions ($\mathcal{D}$) and alternative solutions ($\mathcal{D}'$). The performance gpa is calculated by $\Delta=\mathcal{D}-\mathcal{D}'$. The best-performed results in each group are marked with \underline{underline}.}
\vspace{-0.2cm}
\label{tab:result}
\resizebox{0.96\textwidth}{!}{
\begin{tabular}{@{}ccccccccccccc@{}}
\toprule
\multicolumn{1}{c|}{\multirow{2}{*}{Model}} & \multicolumn{4}{c|}{$\mathcal{D}$} & \multicolumn{4}{c|}{$\mathcal{D}'$} & \multicolumn{4}{c}{$\Delta$} \\ \cmidrule(l){2-13} 
\multicolumn{1}{c|}{} &\ \ $\mathcal{M}_0$\ \ &\ \ $\mathcal{M}_1$\ \ &\ \ $\mathcal{M}_2$\ \ & \multicolumn{1}{c|}{\ \ $\mathcal{M}_3$\ \ } &\ \ $\mathcal{M}_0$\ \ &\ \ $\mathcal{M}_1$\ \ &\ \ $\mathcal{M}_2$\ \ & \multicolumn{1}{c|}{\ \ $\mathcal{M}_3$\ \ } &\ \ $\mathcal{M}_0$\ \ &\ \ $\mathcal{M}_1$\ \ &\ \ $\mathcal{M}_2$\ \ &\ \ $\mathcal{M}_3$\ \ \\ \midrule
 & \multicolumn{12}{c}{Base} \\ \midrule
\multicolumn{1}{c|}{GPT-4o} & 27.2 & 47.7 & 48.8 & \multicolumn{1}{c|}{\underline{59.1}} & 18.4 & 36.7 & 37.8 & \multicolumn{1}{c|}{\underline{48.5}} & \underline{-8.8} & -11.0 & -11.0 & -10.6 \\
\multicolumn{1}{c|}{Claude-3.5} & 38.2 & 51.1 & 50.7 & \multicolumn{1}{c|}{\underline{56.6}} & 34.7 & 48.5 & 44.0 & \multicolumn{1}{c|}{\underline{51.8}} & -3.5 & \underline{-2.6} & -6.7 & -4.8 \\
\multicolumn{1}{c|}{Gemini-1.5} & 46.4 & 54.2 & 61.7 & \multicolumn{1}{c|}{\underline{62.4}} & 39.5 & 49.5 & 55.5 & \multicolumn{1}{c|}{\underline{59.5}} & -6.9 & -4.7 & -6.2 & \underline{-2.9} \\
\multicolumn{1}{c|}{Llama-3.1} & 20.2 & 34.6 & 23.5 & \multicolumn{1}{c|}{\underline{37.9}} & 20.9 & 31.0 & 23.6 & \multicolumn{1}{c|}{\underline{32.4}} & \underline{+0.7} & -3.6 & +0.1 & -5.4 \\
\multicolumn{1}{c|}{Qwen-2.5} & 24.7 & 40.3 & 34.4 & \multicolumn{1}{c|}{\underline{44.3}} & 16.3 & 35.3 & 25.0 & \multicolumn{1}{c|}{\underline{38.3}} & -8.4 & \underline{-5.0} & -9.5 & -6.0 \\ \midrule
 & \multicolumn{12}{c}{Advance} \\ \midrule
\multicolumn{1}{c|}{GPT-4o} & 52.9 & 63.4 & \underline{67.1} & \multicolumn{1}{c|}{66.3} & 43.4 & 59.3 & 58.0 & \multicolumn{1}{c|}{\underline{61.4}} & -9.5 & -4.1 & -9.1 & \underline{-4.9} \\
\multicolumn{1}{c|}{Claude-3.5} & 59.0 & 61.7 & 72.5 & \multicolumn{1}{c|}{\underline{73.1}} & 52.7 & 57.0 & 67.4 & \multicolumn{1}{c|}{\underline{69.5}} & -6.3 & -4.7 & -5.2 & \underline{-3.6} \\
\multicolumn{1}{c|}{Gemini-1.5} & 65.2 & 65.6 & \underline{76.0} & \multicolumn{1}{c|}{71.8} & 55.6 & 58.1 & \underline{72.2} & \multicolumn{1}{c|}{68.8} & -9.6 & -7.5 & -3.8 & \underline{-3.0} \\
\multicolumn{1}{c|}{Llama-3.1} & 44.3 & 64.0 & 63.4 & \multicolumn{1}{c|}{\underline{71.9}} & 37.2 & 56.4 & 57.1 & \multicolumn{1}{c|}{\underline{67.9}} & -7.2 & -7.6 & -6.3 & \underline{-4.0} \\
\multicolumn{1}{c|}{Qwen-2.5} & 46.3 & 57.4 & 45.4 & \multicolumn{1}{c|}{\underline{60.2}} & 38.8 & 50.6 & 43.0 & \multicolumn{1}{c|}{\underline{58.2}} & -7.5 & -6.8 & -2.4 & \underline{-2.0} \\ \bottomrule
\end{tabular}}
\vspace{-0.5cm}
\end{table*}

In Table~\ref{tab:result}, we present the comparison between baseline methods and AskDB over both the conventional solutions $\mathcal{D}$ and alternative solution $\mathcal{D}'$. 
To address RQ1, we analyze the values in the $\Delta$ columns between $\mathcal{M}_2$ and $\mathcal{M}_0$. The table clearly demonstrates that AskDB is effective in mitigating conformity bias in error detection results for advanced versions of LLMs. However, for base LLMs, the benefits of naively applying AskDB are less evident. Among the five LLM frameworks, only Gemini exhibits a reduced performance gap between $\mathcal{D}$ and $\mathcal{D}'$. We attribute this to the relatively weaker reasoning capabilities of base models. With the naive instruction prompt, these models fail to fully leverage the valuable information provided by the reference solutions, thereby limiting the effectiveness of AskDB in these cases. Comparing $\mathcal{M}_1$ with $\mathcal{M}_2$, we observe that the CoT prompt strategy also helps mitigate conformity bias in LLM-powered error detectors. Nevertheless, in most advanced models, AskDB consistently outperforms CoT in narrowing the gap between conventional and alternative solutions.

To answer RQ2, we compare $\mathcal{M}_2$ with $\mathcal{M}_0$ in the $\mathcal{D}$ and $\mathcal{D}'$ columns. The results indicate a consistent improvement in error detection accuracy after adopting the AskDB framework. This suggests that AskDB not only helps reduce the performance gap between conventional and alternative solutions but also enhances overall detection performance. Comparing $\mathcal{M}_1$ with $\mathcal{M}_2$, we find that AskDB and CoT prompts enable different LLMs to achieve better results. In summary, AskDB is more compatible with advanced models, while CoT demonstrates greater efficacy with base-sized models.

To address RQ3, we compare $\mathcal{M}_3$ with $\mathcal{M}_1$. The results reveal that AskDB is highly compatible with other reasoning-enhancing techniques such as CoT prompts in the context of error detection. For advanced model of Llama-3.1 and base model of Gemini-1.5, combining AskDB with CoT yields significant performance improvements compared to using either method independently. These findings confirm that AskDB is a robust approach for mitigating conformity bias. Moreover, its compatibility with other reasoning-enhancement techniques achieves the best overall performance in error detection tasks.

\vspace{-0.2cm}
\section{Related Work}
\vspace{-0.2cm}
\subsection{Automatic Error Detection}

Automatic error detection (AED) is a widely studied research task in the field of education~\cite{zamora2018error}. Since the advent of pre-trained language models (PLMs) such as BERT~\cite{kenton2019bert}, AED algorithms in language education have achieved significant advancements~\cite{bryant2023grammatical}. Applications like grammar error detection have been widely implemented in the teaching of languages~\cite{he2021english}. Moreover, by integrating PLMs with acoustic models, AED has also shown promising results in detecting pronunciation errors~\cite{wei2022automatic}. The recent emergence of large language models (LLMs) has further expanded the scope of AED research beyond language education. Leveraging their advanced capabilities in mathematical reasoning~\cite{ahn2024large}, task planning~\cite{huang2024understanding}, and even programming~\cite{nam2024using}, LLMs have been increasingly adopted in recent studies to develop AED solutions for complex educational subjects, such as programming~\cite{gabbay2024combining} and mathematics~\cite{yan2024errorradar}. 


\vspace{-0.2cm}
\subsection{Math Reasoning in LLMs}
\vspace{-0.2cm}

Reasoning capability is one of the most attractive features reported among the emergent capabilities of large language models (LLMs). Building on approaches such as chain-of-thought~\cite{wei2022chain}, LLMs have demonstrated impressive performance in solving complex logical reasoning problems. However, recent studies~\cite{prabhakar2024deciphering} have raised skepticism about these reasoning capabilities, suggesting they may primarily originate from memorization rather than genuine reasoning ability. To address these concerns, numerous new reasoning tasks and benchmark datasets have been introduced~\cite{srivastava2024functional}. Among these, approaches that involve error detection and correction of flawed solutions have gained popularity in the community as a means to evaluate true mathematical reasoning capabilities, aided by the availability of extensive benchmark datasets~\cite{,srivastava2024functional,co}. To reduce the burden of tedious human annotation, many recent works have proposed algorithms to automatically generate inputs for these tasks based on existing datasets~\cite{li2024evaluating}. Through extensive experiments on these newly introduced mathematical reasoning tasks, the reasoning capabilities of LLMs have been rigorously evaluated and significantly validated. Moreover, with the rapid advancements in multi-modal large language models, investigating the multimodal mathematical reasoning capabilities of current vision-language LLMs is becoming an increasingly prominent area of research~\citep{yan2024errorradar}.

\vspace{-0.2cm}
\section{Conclusion}
\vspace{-0.2cm}
In this work, we investigate the behavior of LLM-powered error detectors when encountering alternative solutions commonly found in real-world math word problems. Through a preliminary study on an alternative solution error detection dataset, we identify and confirm the presence of conformity bias in LLMs during error detection. Motivated by our findings on the impact of incorporating reference solutions, we propose the Ask-Before-Detection (AskBD) framework, which enhances error detection by adaptively generating reference solutions. Comprehensive experiments on 200 examples from GSM8K demonstrate the effectiveness of AskBD in mitigating conformity bias. Furthermore, when combined with reasoning enhancement techniques like chain-of-thought (CoT) prompting, AskBD achieves significant improvements in both bias mitigation and overall performance.

\section*{Limitation}
In this work, we identify conformity bias in LLM-powered error detectors for math word problem (MWP) solutions using 200 seed samples from the GSM8K dataset. During the data preparation process, we selected four common error types in student solutions as targets to simulate real-world error detection scenarios. However, this approach has limitations, as it overlooks the occurrence of rarer but potentially more challenging error types in student solutions. To address this, we plan to collect samples from real-world student responses in future iterations of our study. Additionally, this work focuses exclusively on alternative solutions for math word problems. The phenomenon of multiple valid solutions to a single problem is widespread across other subjects in education. In future research, we aim to extend our analysis of conformity bias to these subjects, contributing to the development of LLM-powered detectors as more effective tools in educational contexts.

\bibliography{reference}

\begin{thebibliography}{28}
\providecommand{\natexlab}[1]{#1}

\bibitem[{Ahn et~al.(2024)Ahn, Verma, Lou, Liu, Zhang, and Yin}]{ahn2024large}
Janice Ahn, Rishu Verma, Renze Lou, Di~Liu, Rui Zhang, and Wenpeng Yin. 2024.
\newblock Large language models for mathematical reasoning: Progresses and
  challenges.
\newblock \emph{arXiv preprint arXiv:2402.00157}.

\bibitem[{Anthropic(2024)}]{TheC3}
Anthropic. 2024.
\newblock \href {https://api.semanticscholar.org/CorpusID:268232499} {The
  claude 3 model family: Opus, sonnet, haiku}.

\bibitem[{Bryant et~al.(2023)Bryant, Yuan, Qorib, Cao, Ng, and
  Briscoe}]{bryant2023grammatical}
Christopher Bryant, Zheng Yuan, Muhammad~Reza Qorib, Hannan Cao, Hwee~Tou Ng,
  and Ted Briscoe. 2023.
\newblock Grammatical error correction: A survey of the state of the art.
\newblock \emph{Computational Linguistics}, 49(3):643--701.

\bibitem[{Bubeck et~al.(2023)Bubeck, Chandrasekaran, Eldan, Gehrke, Horvitz,
  Kamar, Lee, Lee, Li, Lundberg et~al.}]{bubeck2023sparks}
S{\'e}bastien Bubeck, Varun Chandrasekaran, Ronen Eldan, Johannes Gehrke, Eric
  Horvitz, Ece Kamar, Peter Lee, Yin~Tat Lee, Yuanzhi Li, Scott Lundberg,
  et~al. 2023.
\newblock Sparks of artificial general intelligence: Early experiments with
  gpt-4.
\newblock \emph{arXiv preprint arXiv:2303.12712}.

\bibitem[{Cobbe et~al.(2021)Cobbe, Kosaraju, Bavarian, Chen, Jun, Kaiser,
  Plappert, Tworek, Hilton, Nakano et~al.}]{cobbe2021training}
Karl Cobbe, Vineet Kosaraju, Mohammad Bavarian, Mark Chen, Heewoo Jun, Lukasz
  Kaiser, Matthias Plappert, Jerry Tworek, Jacob Hilton, Reiichiro Nakano,
  et~al. 2021.
\newblock Training verifiers to solve math word problems.
\newblock \emph{arXiv preprint arXiv:2110.14168}.

\bibitem[{Daheim et~al.(2024)Daheim, Macina, Kapur, Gurevych, and
  Sachan}]{daheim2024stepwise}
Nico Daheim, Jakub Macina, Manu Kapur, Iryna Gurevych, and Mrinmaya Sachan.
  2024.
\newblock Stepwise verification and remediation of student reasoning errors
  with large language model tutors.
\newblock \emph{arXiv preprint arXiv:2407.09136}.

\bibitem[{Gabbay and Cohen(2024)}]{gabbay2024combining}
Hagit Gabbay and Anat Cohen. 2024.
\newblock Combining llm-generated and test-based feedback in a mooc for
  programming.
\newblock In \emph{Proceedings of the Eleventh ACM Conference on Learning@
  Scale}, pages 177--187.

\bibitem[{He(2021)}]{he2021english}
Zhenhui He. 2021.
\newblock English grammar error detection using recurrent neural networks.
\newblock \emph{Scientific Programming}, 2021(1):7058723.

\bibitem[{Huang et~al.(2023)Huang, Zou, Cheng, Chen, and Xie}]{huang2023trends}
Xinyi Huang, Di~Zou, Gary Cheng, Xieling Chen, and Haoran Xie. 2023.
\newblock Trends, research issues and applications of artificial intelligence
  in language education.
\newblock \emph{Educational Technology \& Society}, 26(1):112--131.

\bibitem[{Huang et~al.(2024)Huang, Liu, Chen, Wang, Wang, Lian, Wang, Tang, and
  Chen}]{huang2024understanding}
Xu~Huang, Weiwen Liu, Xiaolong Chen, Xingmei Wang, Hao Wang, Defu Lian, Yasheng
  Wang, Ruiming Tang, and Enhong Chen. 2024.
\newblock Understanding the planning of llm agents: A survey.
\newblock \emph{arXiv preprint arXiv:2402.02716}.

\bibitem[{Jiang et~al.(2024)Jiang, Peng, Feng, Li, and Li}]{jiang2024llms}
Zhuoxuan Jiang, Haoyuan Peng, Shanshan Feng, Fan Li, and Dongsheng Li. 2024.
\newblock Llms can find mathematical reasoning mistakes by pedagogical
  chain-of-thought.
\newblock \emph{arXiv preprint arXiv:2405.06705}.

\bibitem[{Kenton and Toutanova(2019)}]{kenton2019bert}
Jacob Devlin Ming-Wei~Chang Kenton and Lee~Kristina Toutanova. 2019.
\newblock Bert: Pre-training of deep bidirectional transformers for language
  understanding.
\newblock In \emph{Proceedings of naacL-HLT}, volume~1, page~2. Minneapolis,
  Minnesota.

\bibitem[{Leacock et~al.(2014)Leacock, Chodorow, Gamon, and
  Tetreault}]{leacock2014automated}
Claudia Leacock, Martin Chodorow, Michael Gamon, and Joel Tetreault. 2014.
\newblock \emph{Automated grammatical error detection for language learners}.
\newblock Morgan \& Claypool Publishers.

\bibitem[{Li et~al.(2024)Li, Wang, Li, Guo, Zhang, and Feng}]{li2024evaluating}
Xiaoyuan Li, Wenjie Wang, Moxin Li, Junrong Guo, Yang Zhang, and Fuli Feng.
  2024.
\newblock Evaluating mathematical reasoning of large language models: A focus
  on error identification and correction.
\newblock \emph{arXiv preprint arXiv:2406.00755}.

\bibitem[{Messer et~al.(2024)Messer, Brown, K{\"o}lling, and
  Shi}]{messer2024automated}
Marcus Messer, Neil~CC Brown, Michael K{\"o}lling, and Miaojing Shi. 2024.
\newblock Automated grading and feedback tools for programming education: A
  systematic review.
\newblock \emph{ACM Transactions on Computing Education}, 24(1):1--43.

\bibitem[{Min et~al.(2023)Min, Ross, Sulem, Veyseh, Nguyen, Sainz, Agirre,
  Heintz, and Roth}]{min2023recent}
Bonan Min, Hayley Ross, Elior Sulem, Amir Pouran~Ben Veyseh, Thien~Huu Nguyen,
  Oscar Sainz, Eneko Agirre, Ilana Heintz, and Dan Roth. 2023.
\newblock Recent advances in natural language processing via large pre-trained
  language models: A survey.
\newblock \emph{ACM Computing Surveys}, 56(2):1--40.

\bibitem[{Nam et~al.(2024)Nam, Macvean, Hellendoorn, Vasilescu, and
  Myers}]{nam2024using}
Daye Nam, Andrew Macvean, Vincent Hellendoorn, Bogdan Vasilescu, and Brad
  Myers. 2024.
\newblock Using an llm to help with code understanding.
\newblock In \emph{Proceedings of the IEEE/ACM 46th International Conference on
  Software Engineering}, pages 1--13.

\bibitem[{Pan et~al.(2023)Pan, Albalak, Wang, and Wang}]{pan2023logic}
Liangming Pan, Alon Albalak, Xinyi Wang, and William~Yang Wang. 2023.
\newblock Logic-lm: Empowering large language models with symbolic solvers for
  faithful logical reasoning.
\newblock \emph{arXiv preprint arXiv:2305.12295}.

\bibitem[{Prabhakar et~al.(2024)Prabhakar, Griffiths, and
  McCoy}]{prabhakar2024deciphering}
Akshara Prabhakar, Thomas~L Griffiths, and R~Thomas McCoy. 2024.
\newblock Deciphering the factors influencing the efficacy of chain-of-thought:
  Probability, memorization, and noisy reasoning.
\newblock \emph{arXiv preprint arXiv:2407.01687}.

\bibitem[{Srivastava et~al.(2024)Srivastava, PV, Menon, Sukumar, Philipose,
  Prince, Thomas et~al.}]{srivastava2024functional}
Saurabh Srivastava, Anto PV, Shashank Menon, Ajay Sukumar, Alan Philipose,
  Stevin Prince, Sooraj Thomas, et~al. 2024.
\newblock Functional benchmarks for robust evaluation of reasoning performance,
  and the reasoning gap.
\newblock \emph{arXiv preprint arXiv:2402.19450}.

\bibitem[{Team et~al.(2023)Team, Anil, Borgeaud, Alayrac, Yu, Soricut,
  Schalkwyk, Dai, Hauth, Millican et~al.}]{team2023gemini}
Gemini Team, Rohan Anil, Sebastian Borgeaud, Jean-Baptiste Alayrac, Jiahui Yu,
  Radu Soricut, Johan Schalkwyk, Andrew~M Dai, Anja Hauth, Katie Millican,
  et~al. 2023.
\newblock Gemini: a family of highly capable multimodal models.
\newblock \emph{arXiv preprint arXiv:2312.11805}.

\bibitem[{Touvron et~al.(2023)Touvron, Martin, Stone, Albert, Almahairi,
  Babaei, Bashlykov, Batra, Bhargava, Bhosale et~al.}]{touvron2023llama}
Hugo Touvron, Louis Martin, Kevin Stone, Peter Albert, Amjad Almahairi, Yasmine
  Babaei, Nikolay Bashlykov, Soumya Batra, Prajjwal Bhargava, Shruti Bhosale,
  et~al. 2023.
\newblock Llama 2: Open foundation and fine-tuned chat models.
\newblock \emph{arXiv preprint arXiv:2307.09288}.

\bibitem[{Wei et~al.(2022{\natexlab{a}})Wei, Wang, Schuurmans, Bosma, Xia, Chi,
  Le, Zhou et~al.}]{wei2022chain}
Jason Wei, Xuezhi Wang, Dale Schuurmans, Maarten Bosma, Fei Xia, Ed~Chi, Quoc~V
  Le, Denny Zhou, et~al. 2022{\natexlab{a}}.
\newblock Chain-of-thought prompting elicits reasoning in large language
  models.
\newblock \emph{Advances in neural information processing systems},
  35:24824--24837.

\bibitem[{Wei et~al.(2022{\natexlab{b}})Wei, Cucchiarini, van Hout, and
  Strik}]{wei2022automatic}
Xing Wei, Catia Cucchiarini, Roeland van Hout, and Helmer Strik.
  2022{\natexlab{b}}.
\newblock Automatic speech recognition and pronunciation error detection of
  dutch non-native speech: cumulating speech resources in a pluricentric
  language.
\newblock \emph{Speech Communication}, 144:1--9.

\bibitem[{Yan et~al.(2024)Yan, Wang, Huo, Li, Li, Su, Gao, Zhang, Xu, Chu
  et~al.}]{yan2024errorradar}
Yibo Yan, Shen Wang, Jiahao Huo, Hang Li, Boyan Li, Jiamin Su, Xiong Gao,
  Yi-Fan Zhang, Tianlong Xu, Zhendong Chu, et~al. 2024.
\newblock Errorradar: Benchmarking complex mathematical reasoning of multimodal
  large language models via error detection.
\newblock \emph{arXiv preprint arXiv:2410.04509}.

\bibitem[{Yang et~al.(2024)Yang, Yang, Hui, Zheng, Yu, Zhou, Li, Li, Liu, Huang
  et~al.}]{yang2024qwen2}
An~Yang, Baosong Yang, Binyuan Hui, Bo~Zheng, Bowen Yu, Chang Zhou, Chengpeng
  Li, Chengyuan Li, Dayiheng Liu, Fei Huang, et~al. 2024.
\newblock Qwen2 technical report.
\newblock \emph{arXiv preprint arXiv:2407.10671}.

\bibitem[{Zamora et~al.(2018)Zamora, Su{\'a}rez, and Ardura}]{zamora2018error}
{\'A}ngela Zamora, Jos{\'e}~Manuel Su{\'a}rez, and Diego Ardura. 2018.
\newblock Error detection and self-assessment as mechanisms to promote
  self-regulation of learning among secondary education students.
\newblock \emph{The Journal of Educational Research}, 111(2):175--185.

\bibitem[{Zhou et~al.(2024)Zhou, Liu, Ning, Liu, Wang, Wong, Huang, Wang, and
  Huang}]{zhou2024your}
Zihao Zhou, Shudong Liu, Maizhen Ning, Wei Liu, Jindong Wang, Derek~F Wong,
  Xiaowei Huang, Qiufeng Wang, and Kaizhu Huang. 2024.
\newblock Is your model really a good math reasoner? evaluating mathematical
  reasoning with checklist.
\newblock \emph{arXiv preprint arXiv:2407.08733}.

\end{thebibliography}

\appendix
\section{Dataset Statistics}
\label{ap:data_detail}

\begin{table}[!btph]
\caption{Statistics on conventional solutions ($\mathcal{D}$) and alternative solutions ($\mathcal{D}'$) across different error categories.}
\label{tab:stats}
\resizebox{0.47\textwidth}{!}{
\begin{tabular}{@{}cccccc@{}}
\toprule
\multirow{2}{*}{\ \ Solution \ \ } & \multirow{2}{*}{Correct} & \multicolumn{4}{c}{Error} \\ \cmidrule(l){3-6} 
 &  &\ \ $\mathcal{E}_{C}$ &\ $\mathcal{E}_{I}$\ &\ $\mathcal{E}_{M}$\ &\ $\mathcal{E}_{H}$\ \ \\ \midrule
$\mathcal{D}$ & 200 & 200 & 200 & 200 & 200 \\
$\mathcal{D}'$ & 200 & 200 & 200 & 200 & 200 \\ \bottomrule
\end{tabular}}
\end{table}

\section{LLM Details}
\label{apx:model_detail}

\begin{table}[!btph]
\centering
\caption{Details about LLM implementation in this paper and source file links.}
\label{tab:llm_detail}
\resizebox{0.38\textwidth}{!}{
\begin{tabular}{@{}cc@{}}
\toprule
LLM Name & Model ID \\ \midrule
\multicolumn{2}{c}{Base} \\ \midrule
GPT-4o & gpt-4o-mini-2024-07-18 \\
Claude-3.5 & claude-3-5-haiku-20241022 \\
Gemini-1.5 & Gemini-1.5-Flash-002 \\
Llama-3.1 & Llama-3.1-8B-Instruct \\
Qwen2.5 & Qwen2.5-7B-Instruct \\ \midrule
\multicolumn{2}{c}{Advance} \\ \midrule
GPT-4o & gpt-4o-2024-08-06 \\
Claude-3.5 & claude-3-5-sonnet-20240620 \\
Gemini-1.5 & Gemini-1.5-Pro-002 \\
Llama-3.1 & Llama-3.1-70B-Instruct \\
Qwen2.5 & Qwen2.5-72B-Instruct \\ \bottomrule
\end{tabular}}
\end{table}

\section{Error Definitions}
\label{apx:error_define}

\begin{table*}[!btph]
\caption{Detailed definition to each error. The error location for each example is marked with the {\color{red}red} color. The original solution is presented at the top of the table for reference.}
\label{tab:error}
\resizebox{\textwidth}{!}{
\begin{tabular}{p{0.15\textwidth}|p{0.3\textwidth}|p{0.75\textwidth}}
\toprule
\textbf{Error} & \textbf{Definition} & \textbf{Example} \\ \midrule
- & Correct solution. & Step 1. Each gust blows the leaf forward 5 feet, so 11 gusts will blow it forward $5 \times 11 = 55$ feet. Step 2. Each swirl after a gust blows it back 2 feet, so 11 swirls will blow it back $2 \times 11 = 22$ feet. Step 3. After 11 gusts, the leaf has traveled $55 - 22 = 33$ feet down the sidewalk. \\ \midrule
$\mathcal{E}_C$ & Operands in expressions are correct but an error occurs in the calculated results. & Step 1. Each gust blows the leaf forward 5 feet, so 11 gusts will blow it forward $5 \times 11 = {\color{red}50}$ feet. Step 2. Each swirl after a gust blows it back 2 feet, so 11 swirls will blow it back $2 \times 11 = 22$ feet. Step 3. After 11 gusts, the leaf has traveled $55 - 22 = 33$ feet down the sidewalk. \\ \midrule
$\mathcal{E}_R$ & Expression are incorrectly referencing the question conditions or the results from prior steps. & Step 1. Each gust blows the leaf forward 5 feet, so 10 gusts will blow it forward $5 \times {\color{red}10} = 50$ feet. Step 2. Each swirl after a gust blows it back 2 feet, so 11 swirls will blow it back $2 \times 11 = 22$ feet. Step 3. After 11 gusts, the leaf has traveled $55 - 22 = 33$ feet down the sidewalk. \\ \midrule
$\mathcal{E}_M$ & Operands or expressions in the step that lack of references or support from the question conditions or prior steps. & Step 1. Each swirl after a gust blows it back 2 feet, so 11 swirls will blow it back $2 \times 11 = 22$ feet. Step 2. After 11 gusts, the leaf has traveled ${\color{red}55} - 22 = 33$ feet down the sidewalk. \\ \midrule
$\mathcal{E}_H$ & Statements or operands in the listed expression are fabricated or inconsistent with the question's conditions. & Step 1. Each gust blows the leaf forward 5 feet, so 11 gusts will blow it forward $5 \times 11 = 55$ feet. Step 2. Each swirl after a gust blows it back 2 feet, so 11 swirls will blow it back $2 \times 11 = 22$ feet. Step 3. After 11 gusts, the leaf has traveled $55 - 22 = 33$ feet down the sidewalk. {\color{red}Finally, a wind blew the leaf 10 feet forward, and the leaf traveled $33 + 10 = 43$ feet.} \\ \bottomrule
\end{tabular}}
\end{table*}

\section{Naive Instruction Prompt}
\label{apx:naive_promot}

\begin{figure*}[!btph]
\centering
\begin{tikzpicture}
  \draw node[draw=black,fill=black!20,rounded corners,inner sep=2ex,text width=0.95\textwidth] {
    Given the <question>, please judge whether each step in <solution> is correct. \textbf{During the judging process,you should know that the <question> does not always have only one standard solution, and any reasonable <solution> should be accepted. You should pay attention to both the expressions and the statements in each step, and take care about the logic consistency between different steps. Additionally, consider arithmetic expression equivalency and avoid rejecting solutions solely because they use equivalent expressions.} 
    \newline

    \textit{In each step, if no errors are found, respond with Step X: <correct>. If you find that the operands in the listed expressions are correct but an error occurs in the calculated result, respond with Step X: <calculation error>. If you find statements or operands in the listed expression are incorrectly referencing the question conditions or the results from prior steps, respond with Step X: <reference error>. If you find operands or expressions in the step that is lack of references or support from the question conditions or prior steps, respond with Step X: <missing step>. If you find statements or operands in the listed expression are fabricated or inconsistent with the question's conditions, respond with: Step X: <hallucination>. If an error is a follow-on issue due to mistakes in previous steps rather than an independent error, respond with: Step X: <secondary error>.}
    \newline \newline
    <question> [Question Text] <solution> [Solution Text]

Now, please start to respond.
    };
\end{tikzpicture}
\caption{The example prompt we used to implement the error detector with LLMs includes specific formatting for clarity. Instruction text guiding the LLMs to accept alternative solutions is highlighted in \textbf{bold}, while the definitions of error categories are emphasized in \textit{italic}. Text enclosed in square brackets serves as placeholders for the input question and solution text, respectively.}
\label{fig:prompt_0}
\end{figure*}

\section{COT Instruction Prompt}
\label{apx:naive_promot}

\begin{figure*}[!btph]
\centering
\begin{tikzpicture}
  \draw node[draw=black,fill=black!20,rounded corners,inner sep=2ex,text width=0.95\textwidth] {
    Given the <question>, please judge whether each step in <solution> is correct. \textbf{During the judging process,you should know that the <question> does not always have only one standard solution, and any reasonable <solution> should be accepted. You should pay attention to both the expressions and the statements in each step, and take care about the logic consistency between different steps. Additionally, consider arithmetic expression equivalency and avoid rejecting solutions solely because they use equivalent expressions.} 
    \newline

    \textit{In each step, if no errors are found, respond with Step X: <correct>. If you find that the operands in the listed expressions are correct but an error occurs in the calculated result, respond with Step X: <calculation error>. If you find statements or operands in the listed expression are incorrectly referencing the question conditions or the results from prior steps, respond with Step X: <reference error>. If you find operands or expressions in the step that is lack of references or support from the question conditions or prior steps, respond with Step X: <missing step>. If you find statements or operands in the listed expression are fabricated or inconsistent with the question's conditions, respond with: Step X: <hallucination>. If an error is a follow-on issue due to mistakes in previous steps rather than an independent error, respond with: Step X: <secondary error>.}
    \newline \newline

    \textbf{\textit{Before the <response>, you should provide your step-by-step <thinking> about your judging process.}} \newline \newline
    <question> [Question Text] <solution> [Solution Text]

Now, please start to \textbf{\textit{think first and then}} respond.
    };
\end{tikzpicture}
\caption{The example CoT prompt we used to implement the error detector with LLMs includes specific formatting for clarity. Instruction text guiding the LLMs to accept alternative solutions is highlighted in \textbf{bold}, while the definitions of error categories are emphasized in \textit{italic}. Text enclosed in square brackets serves as placeholders for the input question and solution text, respectively.}
\label{fig:prompt_1}
\end{figure*}

\end{document}